\documentclass[10pt,twocolumn,letterpaper]{article}

\usepackage{wacv}
\usepackage{times}
\usepackage{epsfig}
\usepackage{graphicx}
\usepackage{amsmath}
\usepackage{amssymb}
\usepackage{booktabs}
\usepackage{authblk}
\usepackage[accsupp]{axessibility}  % Improves PDF readability for those with disabilities.
\def\wacvPaperID{66} % Enter the WACV Paper ID here

\wacvapplicationstrack % Uncomment this line if you are submitting to the Applications Track.

\wacvfinalcopy % *** Uncomment this line for the final submission

\ifwacvfinal
\usepackage[breaklinks=true,bookmarks=false]{hyperref}
\else
\usepackage[pagebackref=true,breaklinks=true,colorlinks,bookmarks=false]{hyperref}
\fi

\begin{document}

%%%%%%%%% TITLE
\title{Semi-Supervised Learning for Sparsely-Labeled Sequential Data: \\
Application to Healthcare Video Processing}

%\author{Florian Dubost\\
%Stanford University,USA\\
%{\tt\small floriandubost1@gmail.com}
%\and
%Erin Hong\\
%{\tt\small erin.hong17@gmail.com}
%}

\author[1]{Florian Dubost*}
\author[1]{Erin Hong*}
\author[1]{Siyi Tang\textsuperscript{1} \protect\\ Nandita Bhaskhar}
\author[1]{Christopher Lee-Messer}
\author[1]{Daniel Rubin}
\affil[1]{Stanford University \authorcr {\tt\small \{floriandubost1, erin.hong17\}@gmail.com, \{cleemess, rubin\}@stanford.edu}}
\affil[*]{{\tt\small equal contribution}}

\maketitle
\thispagestyle{empty}

%%%%%%%%% ABSTRACT
\begin{abstract}
   Labeled data is a critical resource for training and evaluating machine learning models. However, many real-life datasets are only partially labeled. We propose a semi-supervised machine learning training strategy to improve event detection performance on sequential data, such as video recordings, when only sparse labels are available, such as event start times without their corresponding end times.
   Our method uses noisy guesses of the events' end times to train event detection models. Depending on how conservative these guesses are, mislabeled samples may be introduced into the training set.
   We further propose a mathematical model for explaining and estimating the evolution of the classification performance for increasingly noisier end time estimates. We show that neural networks can improve their detection performance by leveraging more training data with less conservative approximations despite the higher proportion of incorrect labels. 
   We adapt sequential versions of CIFAR-10 and MNIST, and use the Berkeley MHAD and HMBD51 video datasets to empirically evaluate our method, and find that our risk-tolerant strategy outperforms conservative estimates by 3.5 points of mean average precision for CIFAR, 30 points for MNIST, 3 points for MHAD, and 14 points for HMBD51.
   Then, we leverage the proposed training strategy to tackle a real-life application: processing continuous video recordings of epilepsy patients, and show that our method outperforms baseline labeling methods by 17 points of average precision, and reaches a classification performance similar to that of fully supervised models. We share part of the code for this article at the following repository:  \href{https://github.com/fpgdubost/CIFAR-10-Sparsely-Labeled-Sequential-Data}{fpgdubost/CIFAR-10-Sparsely-Labeled-Sequential-Data} .
\end{abstract}

%%%%%%%%% BODY TEXT
\section{Introduction}

\label{sec:intro}

\begin{figure*}[t!]
\centering
\includegraphics[width=10cm]{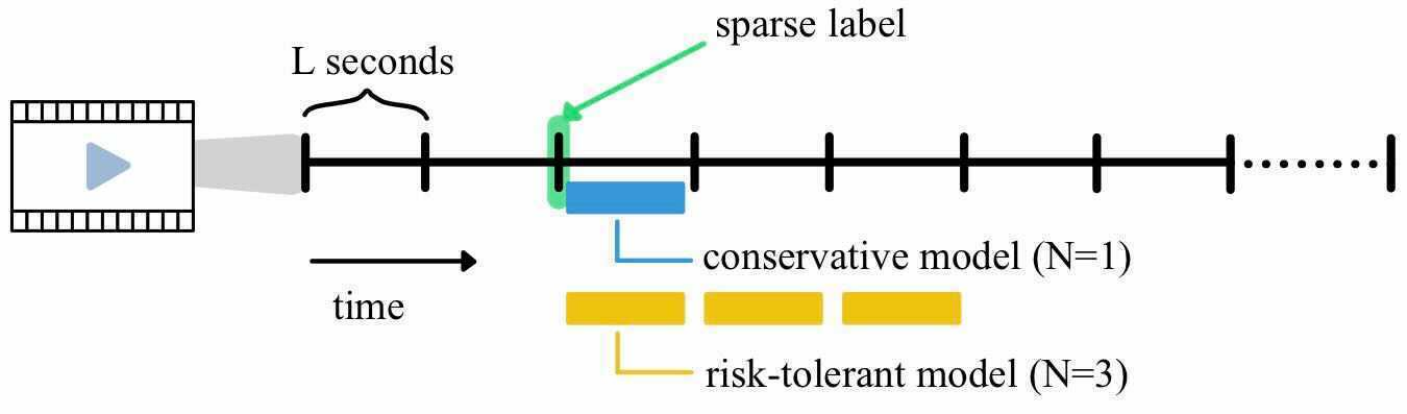}
\caption{
\textbf{Leveraging sparse video labels.} During training time, only event start times are annotated. Event end times have to be guessed, which determine the number $N$ of elements that can be used as positives during training. A \textit{conservative} model only uses a single element ($N=1$) as positive per sparse label, while a \textit{risk-tolerant} model uses multiple elements, e.g. $N=3$. Higher-risk labeling strategies may result in more incorrect labels during training (negative segments being mislabeled as positives). However, these higher-risk strategies can provide more training data, which may result in better detection performance despite training with incorrect labels. 
}
\label{fig:sampling}
\end{figure*}

Labeled image and video datasets are crucial for training and evaluating machine learning models. As a result, computer vision researchers have compiled a number of labeled benchmark datasets, such as MNIST \cite{lecun-mnisthandwrittendigit-2010}, ImageNet \cite{deng2009}, MSCOCO \cite{lin2014}, Kinetics \cite{kay2017}, CIFAR \cite{krizhevsky2009}, and Cityscapes \cite{cordts2016}. However, many application areas still remain poorly covered, such as medical imaging data, despite recent initiatives such as the UK Biobank \cite{sudlow2015}. Although medical institutions often possess large amounts of data, most of it remains unlabeled and underutilized. For example, for research purposes, some hospitals record hours of videos of patients in intensive care, but those videos remain only poorly labeled in the clinical routine, with at best, the sparse event labels.
 
Weakly-supervised learning aims to leverage datasets with either incomplete or incorrect labels. Zhou et al. \cite{zhou2018} identified two subtypes of weak supervision schemes: incomplete and inaccurate supervision.
Incomplete supervision applies when only a portion of the training samples are labeled. For example, semi-supervised learning methods are designed to leverage unlabeled samples next to labeled samples.
Inaccurate supervision applies when the given labels are not necessarily correct (e.g., crowd-sourcing~\cite{maier2014,cheplygina2016}). 
The works of Hao et al. \cite{hao2020} on mammograms and Karimi et al. \cite{karimi2020} on brain MRIs are also examples of inaccurate supervision with deep learning for medical data.  

In this work, we propose a method which combines semi-supervised learning and inaccurate supervision to leverage sparsely-labeled sequential data.
The main task is to detect sequences of events, given only sparse training labels, i.e., the start times of these events. The end times and the duration of these events remain unknown, which prevents sampling any positives events with certainty (Figure \ref{fig:sampling}). For example, in a cooking videos dataset, sparse training labels could indicate when cooking an ingredient started at time T, but without any information about when the cooking of that ingredient stopped. 

To address this problem, we propose making a noisy approximation of event end times. For each sparse label, we choose a fixed number of consecutive elements in sequence that follow the sparse label, and use them as positive training samples (essentially providing a noisy estimate of duration). In the above example with cooking videos, we could guess that the cooking of the ingredient lasts one minute, or 1500 frames at 25 frames per second, and use all 1500 frames as positive samples. The longer the estimated guess, the more likely it is that we introduce potential incorrectly-labeled samples (false positives in the training set).

We further propose a mathematical model for explaining and estimating the evolution of the classification performance for increasingly noisier end time estimates. This model include two sigmoid-based components respectively describing the positive and negative impact of additional noisy labeled sequence elements on the performance.

We empirically evaluate our method on sparsely-labeled sequences of CIFAR-10 images, MNIST images, Berkeley MHAD videos, and HMDB51, and show an improvement of 3.5 points of mean average precision for CIFAR, 30 points for MNIST, 3 points for MHAD, and 14 points for HMDB51 over the baseline method.

Finally, we demonstrate our method on a real-life sequential analysis task---video monitoring of epileptic hospital patients.
Electroencephalography (EEG) is a common modality for recording brain activity and monitoring patients.
Automated methods have been developed to automatically detect seizures from EEG activity \cite{fergus2015,saab2020} but can fail to discern seizures from artifacts caused by disturbances in EEG measurement (e.g., patting on the back or rocking neonatal patients can trigger false positive seizure detections).

We address EEG artifacts by automatically detecting five artifact--suctioning of neonates, chewing, rocking, cares by nurse and patting of neonates--from continuous video recordings acquired during clinical routine.
Those events are annotated with sparse labels (only start times, no end times), which is common practice for the labeling of continuous recordings in clinical routine \cite{siebig2010}.
Our method for learning from sparsely-labeled sequences can leverage those sparse labels, outperforming baseline methods by 17 points of mean average precision. We show that our semi-supervised model can reach the classification of fully supervised models. We also give insight into estimating the parameters of the proposed model in case of merging classes.

To summarize, our main contributions are:
\begin{itemize}
\itemsep0em 
    \item A training strategy for semi-supervised learning with sparsely-labeled sequential data.
    \item A mathematical model for explaining and estimating the evolution of the classification performance for increasingly noisier end time estimates.
    \item A method that automatically detects events from sparsely-labeled continuous video recordings of hospital neonates.
\end{itemize}

\section{Related works}
\label{sec:related}

%\paragraph{Semi-Supervised Learning.}
Semi-supervised training strategies have been developed for a myriad of computer vision tasks. 
In image classification, most state-of-the-art semi-supervised methods are based on self-supervision and use contrastive learning approaches \cite{he2020,chen2020}. 
MoCo \cite{he2020} encodes and matches query images to keys of a dynamic dictionary. SimCLR \cite{chen2020} improves upon MoCo by removing the need of specialised architecture. The authors of SimCLR claimed that the composition of data augmentation is crucial in achieving a high performance.
% Another of their key findings is that high batch sizes substantially improved the performance of contrastive learning.
Earlier, MixMatch \cite{berthelot2019} had already advocated the importance of data augmentation for semi-supervised learning for image classification. Given unlabeled images, MixMatch generated a set of augmented images, passed the images through the network and guessed the label using the mean of the model's predictions. The distribution of predictions was then sharpened using the guessed label.

Semi-supervised learning has also been proposed for semantic image segmentation \cite{bortsova2019,chatterjee2020}. Bortsova et al \cite{bortsova2019} proposed augmenting unlabeled images, and applying reverse transformations to the output segmentations. The transformed segmentations were then forced to be similar to the originals using a consistency loss term.

Jing et al. \cite{jing2021} proposed a semi-supervised learning method for video classification, using pseudo-labels and normalization probabilities of unlabeled videos to improve the classification performance.
% Despite the fact that they are solving a segmentation problem,
% We believe the work of  to be the closest to ours. 
Sibechi et al. \cite{sibechi2019} proposed a semi-supervised method for the segmentation of sparsely-labeled video data. Our work leverages the sequentiality of the data through sampling, while Sibechi et al. leverages the sequentiality by directly including it as a architectural component of their model. Overall, sparsely-labeled video data is less studied than imaging data, which opens an avenue for novel semi-supervised learning methods that can leverage the sequentiality of frames.

 Most standard semi-supervised methods disregard the sequentiality of labels by considering samples i.i.d.. In this article, we leverage the sequentiality of the data by estimating event end times from start times.

\section{Method}
\subsection{Training Strategy}
\label{sec:method}
We consider an infinite sequence $(x_{n})_{n\in \mathbb{N}}$, and a sparse label $l$ indicating the start of an event, i.e., a subsequence of consecutive positive elements $(x_{n})_{n\in [l,l+M-1]}$, with $M$ being the duration of the event. The remaining elements outside of this positive subsequence are considered negative elements.
Typically, if the length of positive subsequence, $M$ is known, we can create a sequence of corresponding binary labels $(y_{n})_{n\in \mathbb{N}}$, where $y_{n} = 1$ if $n \in [l,l+M-1]$ and $y_{n} = 0$ otherwise. Together, both sequences $(x_{n})_{n\in \mathbb{N}}$ and $(y_{n})_{n\in \mathbb{N}}$ can be used to train a machine learning model to detect positive events.
 Multiclass classificaton, with events from multiple classes, will be reviewed later in the method section.

In this work, we would like to train such models under the conditions that the length $M$ is unknown and only sparse labels $l$ indicating the start of the subsequence are known during training time.

To address this problem, we propose making a noisy estimate of $M$, using parameter $N$, which in turn sets the \textit{risk} associated with this estimate. We assign $N$ elements  $(x_{n})_{n\in [l,l+N-1]}$ that follow the labeled positive element $x_{l}$ to be positives. This results in potentially inaccurate labels $(\hat{y}_{n})_{n\in \mathbb{N}}$, with $\hat{y}_{n} = 1$ if $n \in [l,l+N-1]$ and $\hat{y}_{n} = 0$ otherwise. 
 
When $N \leq M$, the $N$ selected elements $\{x_{l},...,x_{l+N-1}\}$ are true positives, and none of the training elements are mislabeled. But when $N<M$, our estimate misses $M-N$ true positive elements $\{x_{l+N},...,x_{l+M-1}\}$ that will not be used for training. This can be suboptimal, especially in datasets where positives are rare. On the other hand, when $N>M$, this method introduces $N-M$ false positive elements $\{x_{l+M},...,x_{l+N-1}\}$ that are incorrectly labeled as positive for training. When $N \geq M$, the higher the value of $N$ is, the higher the proportion of incorrectly labeled samples in the training set and higher the risk.
 
Sampling negative subsequences is simpler. Negatives can be sampled in sequences that do not have positive sparse labels.  In addition, assuming only one positive event exists in the sequence, negative training elements can also be safely sampled before the labeled time $l$ as $(x_{n})_{n\in [0,l-1]}$, because $(y_{n})_{n\in [0,l-1]} = (\hat{y}_{n})_{n\in [0,l-1]}$. Other negative elements can be reasonably safely sampled far away from the labeled time as $(x_{n})_{n\in [P,\infty]}$, provided $P \gg N$.

We train neural networks using sets of incorrectly labeled sequences, with a fixed risk level $N$ for all sequences, and evaluate the detection performance on independent sequences where the length $M$ is known.

\subsection{Impact on Classification Performance}

The number elements $N$ sampled after the sparse labels can have a double-sided impact on the classification performance. As long as $N<M$, an increasing $N$ is likely to improve the classification performance by improving the recall. When $N>M$, we introduce mislabeled training samples that are likely to increase the number of false positive detections. Because of this double-sided impact, we call $N$ the \textit{risk level}.

The positive impact of the number of samples on the classification performance has been shown to follow exponential trends \cite{figueroa2012}, while Brodley et al.'s experiments \cite{brodley1999} showed that the degree of mislabeled data could decrease accuracy according to exponential trends as well. We hypothesize that those separate phenomenons can each be modelled using sigmoid functions, and that--following a probabilistic approach--their simultaneous occurrence can be modelled as the product of their individual probability distributions.
Consequently, we propose to model the impact of $N$ on the classification performance as the product of two sigmoid functions:
\begin{equation}
Performance(N) = \frac{1}{1+e^{-\alpha_{1}(N-\beta_{1})}} . \frac{1}{1+e^{\alpha_{2}(N-\beta_{2})}},
\label{eq:perf}
\end{equation}
where the first sigmoid ($\alpha_{1}$ and $\beta_{1}$) models the positive impact of $N$ on the classification performance, and the second sigmoid ($\alpha_{2}$ and $\beta_{2}$) models its negative impact. The parameters $\alpha_{1}$, $\alpha_{2}$, $\beta_{1}$ and $\beta_{2}$ are positive and $\beta_{1}<M<\beta_{2}$.

As illustrated in Figure \ref{fig:model}, for a given classification task and dataset, the choice of $N$ depends on model's parameters  $\alpha_{1}, \beta_{1}, \alpha_{2}$, and $\beta_{2}$. Sometimes the choice of $N$ is critical, i.e for $\alpha_{1} =5, \beta_{1}=5, \alpha_{2} =1$, and $\beta_{2}=10$, while in other cases, $N$ has little influence on the. classification performance for a large range of values, i.e. with $\alpha_{1} =3, \beta_{1}=1, \alpha_{2} =3$, and $\beta_{2}=9$.

\subsection{Multiclass classification}
In binary classification problems, mislabeled elements automatically belong to the other class, and consequently harm the classification performance. This is more complex for multiclass classification, where mislabeled elements do not necessarily impact the classification performance.

For a single-label multiclass classification problem with $C$ classes, we consider that the labels $(y_{n})_{n\in \mathbb{N}}$ not only take their values in $C$ but also in $C_{o}$. $C_{o}$ is a set of classes not included in our classification problem (but present in the data), such that $C \cap C_{o} = \emptyset $ and that $(y_{n})_{n\in \mathbb{N}}$ takes its values in $C \cup C_{o}$. With our sampling strategy, mislabeled elements which labels are in $C_{o}$ do not have a negative impact on the classification performance for the target multiclass problem (including only the classes $C$). 

If all mislabeled elements have their labels in $C_{o}$, the impact of the risk level $N$ in Equation \ref{eq:perf} can be modeled as a single sigmoid:
\begin{equation}
Performance(N) = \frac{1}{1+e^{-\alpha_{1}(N-\beta_{1})}}. 
\label{eq:single_sig}
\end{equation}
While this is unlikely to occur for the overall classification performance, it could occur when inspecting class-wise classification performance. This could indicate that the target sequence is the subset of a larger sequence ignored in the classification problem and which is semantically distinct from the other sequences. Such examples are presented in the hospital video experiments.

To estimate the impact of mislabeled elements on the classification performance, one can estimate the proportion of mislabeled elements which labels are among the target classes $C$. We note $X_m$ the set of all mislabeled elements, and $X_{C_{o}}$ the set of mislabeled elements which labels are in $C_{o}$. The risk of negatively impacting the target classification performance can be computed as: $P_{multi} = 1 - |X_m \cap X_{C_{o}}| / |X_m \cup X_{C_{o}}|$. This risk can also be computed class-wise. We note $X_{m,c}$ the set of all mislabeled elements during the positive sampling for class $c$, and $X_{C_{o},c}$ the subset of those elements which have labels in $C_{o}$. The risk of negative impact associated to the sampling of class $c$ can be computed as: 

\begin{equation}
P_{multi,c} = 1 - |X_{m,c} \cap X_{C_{o},c}| / |X_{m,c} \cup X_{C_{o},c}|. 
\label{eq:overlap}
\end{equation}

\subsection{Network Architectures}
For the experiments on images, and the MHAD dataset, we use 2D convolutional neural networks that take a 2D matrix as input, and output a single logit for binary classification. The architecture is adopted from a small ResNet \cite{he2016}--two 3×3 convolutional layers, followed by a 2×2 max-pooling layer, again two 3×3 convolutional layers, a global average pooling layer, and a fully connected layer followed by a sigmoid activation function, combining the contribution of the different features into a single output in $[0,1]$. The first two convolutional layers has 32 filters each, and the last two convolutional layers, 64 filters each. The convolutions are zero-padded and followed by ReLU activations. We use skip connections between the input and output of two successive convolutional layers. For the experiments on hospital video clip classification, we use a 3D 18 layers ResNet \cite{tran2018} pretrained on Kinetics-400 \cite{kay2017}.

\subsection{Conservative versus Risk-tolerant Models}
We call \textit{conservative} a model trained using the original sparse labeling, i.e. using only the first element following the sparse labels $l$, which correspond to using a risk level $N=1$. We consider this model to be the baseline. We call \textit{risk-tolerant} a model trained using risk-tolerant labeling, i.e. using more than the first element following the sparse labels $l$. In the MNIST and CIFAR experiments, we experiment with risk levels in $[1,9]$. In the video experiments, the risk-tolerant model is trained using a risk level $N=3$. The architecture, initialization and optimizer are the same for all models.

\subsection{Pseudo Labeling Baseline}

Pseudo labeling has been proposed as a simple and efficient semi-supervised learning method \cite{lee2013pseudo} that indirectly leverages entropy minimization \cite{grandvalet2005}. For the experiments on hospital videos, we compared the risk tolerant models to a pseudo labeling approach. Predictions of the risk-conservative model are used to create pseudo labels for as many training samples as used by the risk tolerant model $N=100$, which correspond to one of the risk level $N$ achieving the highest performance on the set set. Subsequently, a new model is trained using these labels.

\begin{figure}[t!]
\centering
\includegraphics[width=7cm]{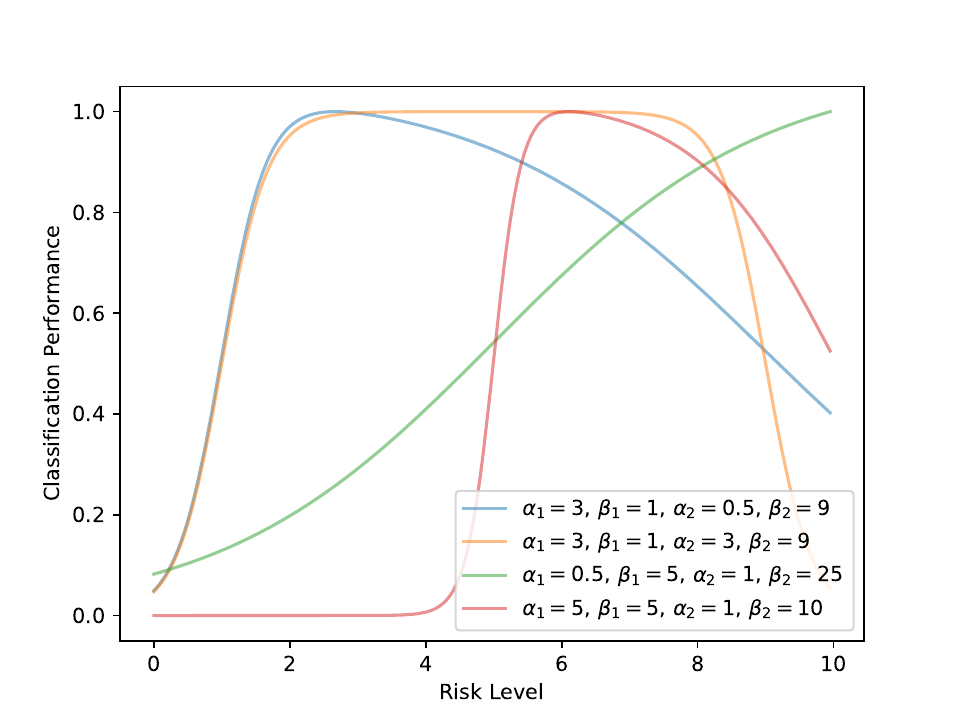}
\caption{
Model of the impact of the risk level $N$ on the classification performance with different sets of parameters. Each curve is rescaled to have its maximum equal to one. The optimal expected value for the risk level $N$ depends on the model's parameters.
}
\label{fig:model}
\end{figure}

\section{Experiments on CIFAR Image Sequences}

To study the proposed training strategy in a controlled setting, we create a toy dataset using CIFAR-10 images.
We find that risk-tolerant labeling outperforms sparse labeling by up to 3.5 points of mean average precision.

\begin{figure*}[t!]
\centering
\includegraphics[width=14cm]{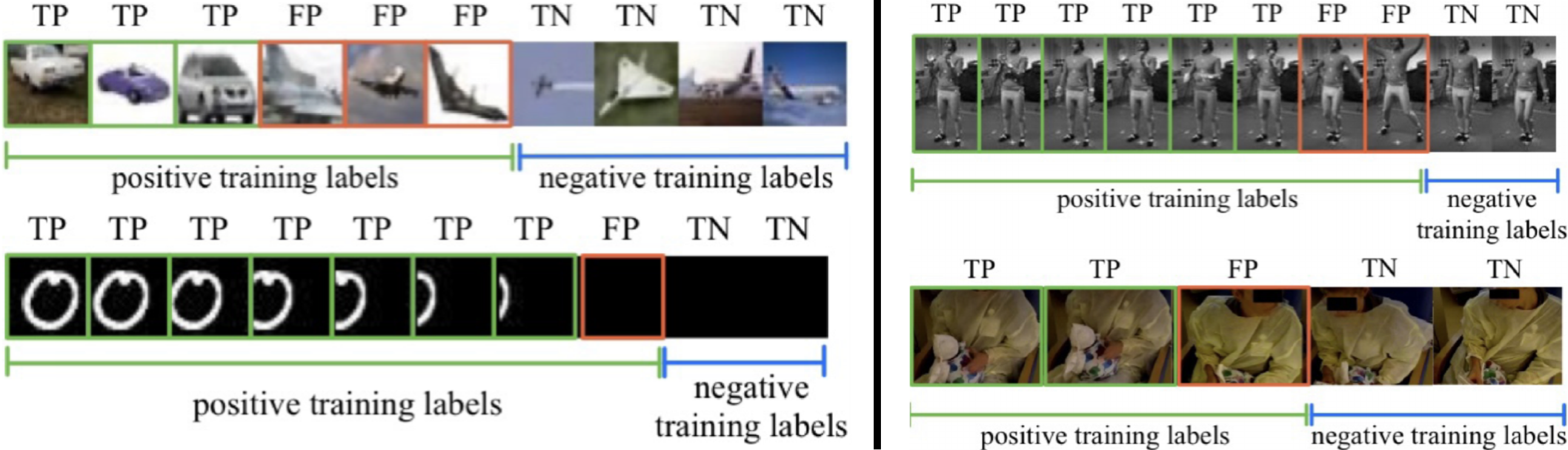}
\caption{
Sequences of automobile vs. airplane CIFAR images (first row), of MNIST images (second row), of Berkeley MHAD video frames (third row), and consecutive hospital video frames (fourth row). Fro example, on the first row, the risk level is $N = 6$, and the true length of the positive subsequence is $M = 3$, which results in $N-M=3$ incorrectly labeled training samples (FP). TP indicates true positives, FP false positives and TN true negatives.}
\label{fig:image_sequences}
\end{figure*}

Experiments on image sequences are designed according to the following scheme. One of the classes is selected as positive -- the automobile (car) class in our experiments -- and another class as negative -- e.g. the airplane class. All images of the training set are equally split into training and validation sets, and the testing images are kept aside. We arrange the training images into 50 sequences of 10 images each by drawing images at random from the two target classes. Each sequence is parameterized by an integer duration $M$ (drawn uniformly from $0 \leq M \leq 10$) of true positive elements, with the first $M$ images drawn from the positive class and the remaining $10-M$ images drawn from the negative class (Figure \ref{fig:image_sequences}).
As we want to model a scenario where the duration $M$ is not known during all the optimization steps, sequence creation is performed similarly for the validation set.

To construct our risk-dependent training labels, we select a risk level $1 \leq N \leq 9$ over all 50 sequences and sample the $N$ first images of each sequence as positives, independent of their true class. If $N>M$, this results in using $N-M$ incorrectly labeled images per sequence for training (false positives in Figure \ref{fig:image_sequences}). Next to this, 50 negative training images are directly sampled from the negative class. For preprocessing, image intensity values are rescaled in $[0,1]$ using the the image-wise minimum and maximum to facilitate the training.

We use this dataset to train a convolutional network with the Adadelta optimizer \cite{zeiler2012} and optimize the binary cross-entropy. For every epoch, the same proportion of negative and positive training images are shown to the network to avoid rebalancing the loss function. Training is stopped after the validation loss diverges, and the best model is selected as the one minimizing the validation loss.
The trained model's performance is evaluated on the left-out test set and measured using recall, precision, F1-score, average precision and AUC. 

All experiments described above are repeated using varying levels of risk $N \in [1,9]$.

The automobile class is chosen as the positive class, and 9 series of experiments are realized using the 9 other classes as negatives, respectively. In the first series of experiments, we create sequences of automobiles and airplanes (Figure \ref{fig:image_sequences}). In the second series of experiments, we create sequences of automobiles and birds, and so forth. 
Each experiment is repeated 10 times using different random initializations of the network's weights. This results in a total of 9 risk levels times 9 classes times 10 runs equals 810 experiments. Detailed results are shown in Table \ref{tab:results_ap_cifar}, Figure \ref{fig:precision} and in supplementary materials.
Overall, risk-tolerant labeling outperforms conservative labeling ($N=1$) by up to 3.5 points of mean average precision with risk level $N=6$. The impact of $N$ on the classification performance in Figure \ref{fig:precision} can be compared to the model with parameters $\alpha_{1} =3, \beta_{1}=1, \alpha_{2} =0.5$, and $\beta_{2}=9$ in Figure \ref{fig:model}. This indicate an early positive contribution of $N$ with a late and progressively increasing negative contribution. In that case, $N$ is better sampled close to the average expected true sequence length $M=5$. Class-wise results in Table \ref{tab:results_ap_cifar} indicate that for most classes, the peak performance is reached for value of $N$ falling between $[3,7]$, i.e. 50\% to 125\% of the average true sequence length $M=5$.

\begin{figure}[t!]
\centering
\includegraphics[width=5cm]{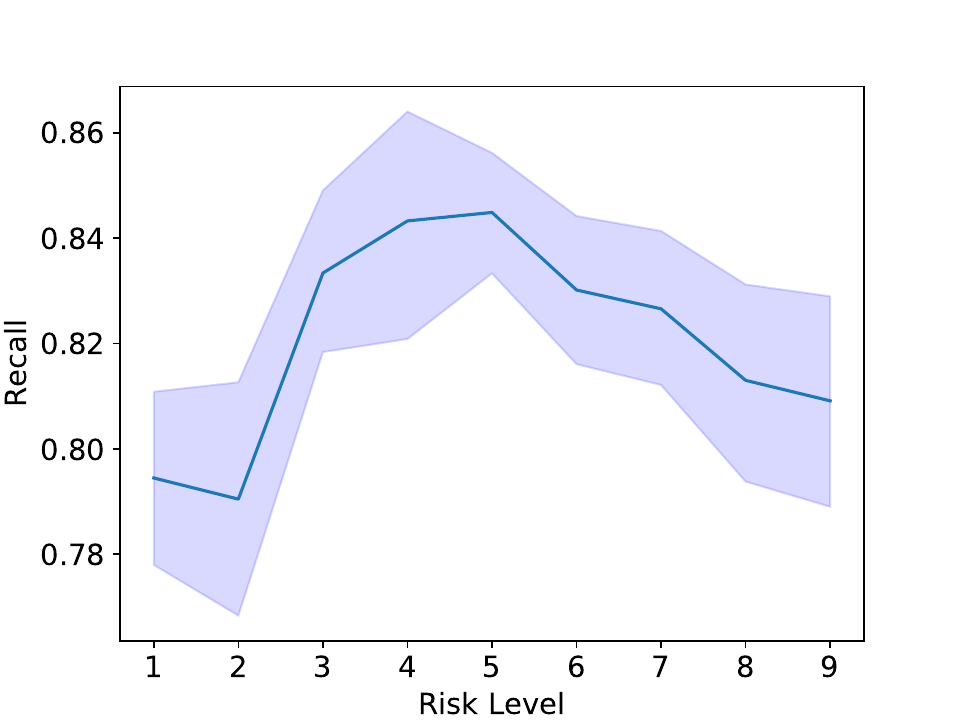}
\caption{
Recall averaged over all 810 the CIFAR-10 experiments, with varying risk levels $N$. 95\% confidence interval are computed with bootstrapping. The baseline method, which does not leverage unlabeled data, is for risk $N=1$.
}
\label{fig:precision}
\end{figure}

\begin{table*}[]
\center
\resizebox{14cm}{!}{%
\begin{tabular}{c|c|c|c|c|c|c|c|c|c|}
Task:    & \multicolumn{9}{c|}{Risk Level (N)}                                                                                                                                                         \\ \cline{2-10} 
car vs.  & 1          & 2          & 3                   & 4                   & 5          & 6                   & 7                   & 8          & 9                   \\ \hline
airplane & 67.7 (1.4) & 70.8 (4.0) & \textbf{77.5 (1.2)} & 74.6 (2.1)          & 71.9 (2.9) & 74.5 (1.3)          & 73.8 (1.5)          & 71.7 (3.3) & 65.8 (4.0)          \\
bird     & 68.8 (2.0) & 60.1 (5.4) & 66.6 (1.7)          & 65.8 (1.4)          & 67.6 (2.0) & 64.8 (1.5)          & 69.1 (2.3)          & 65.6 (0.9) & 67.4 (1.3)          \\
cat      & 56.0 (0.8) & 55.1 (0.9) & 61.3 (1.9)          & 63.1 (1.9)          & 61.2 (1.3) & \textbf{63.6 (1.5)} & 55.2 (1.0)          & 56.2 (1.7) & 57.5 (2.2)          \\
deer     & 60.5 (3.2) & 62.9 (3.1) & 59.8 (1.9)          & \textbf{64.6 (2.0)} & 61.8 (1.2) & 59.5 (0.8)          & 59.1 (1.4)          & 62.1 (1.0) & 62.0 (1.0)          \\
dog      & 59.3 (1.1) & 56.9 (1.5) & \textbf{61.5 (1.7)} & 58.9 (1.0)          & 58.2 (1.6) & 59.6 (2.6)          & 57.0 (1.7)          & 60.6 (1.1) & 60.2 (2.2)          \\
frog     & 50.0 (1.3) & 51.9 (0.6) & 56.8 (2.6)          & 53.5 (2.0)          & 53.1 (1.7) & \textbf{59.5 (3.2)} & 53.1 (0.7)          & 51.2 (1.3) & 49.2 (0.8)          \\
horse    & 55.4 (1.9) & 52.4 (1.4) & 54.0 (0.8)          & 55.7 (2.6)          & 53.0 (1.5) & 60.8 (4.3)          & \textbf{66.0 (2.1)} & 63.9 (2.6) & 55.4 (1.7)          \\
ship     & 66.9 (1.1) & 67.3 (1.3) & \textbf{72.2 (1.6)} & 70.5 (0.7)          & 71.4 (0.9) & 72.3 (2.6)          & 69.8 (1.5)          & 67.6 (1.3) & 69.6 (0.5)          \\
truck    & 55.7 (0.8) & 62.8 (4.7) & 56.0 (2.3)          & 54.4 (4.9)          & 51.5 (2.0) & 60.9 (2.9)          & 47.9 (0.4)          & 58.7 (2.2) & \textbf{63.9 (1.4)}
\end{tabular}
}
\caption{\label{tab:results_ap_cifar}Average Precision for the CIFAR-10 dataset. Means (standard deviations) are computed for 10 repetitions of the experiments with different random initialization of the weights. Best statistically significant (p-value $<$ 0.05) results are highlighted in bold. The baseline method, which does not leverage unlabeled data, is shown in the first column.}
\end{table*}

\section{Experiments on Moving MNIST Digits}
We design a sequential MNIST dataset, in which we incorporate motion and a notion of sequence in our by shifting our subject (digit zero in our experiments) to the left out of the frame of view. The shifting speed is drawn at random for each sequence, and the remaining images are left empty (Figure~\ref{fig:image_sequences}). A image of the sequence is considered true positive if the subject still appears on the image. To complexify the task, Gaussian noise--with a mean of 1 and standard deviation of 2--is added to the images. 

\begin{figure}[t]
\begin{center}
   \includegraphics[width=8.5cm]{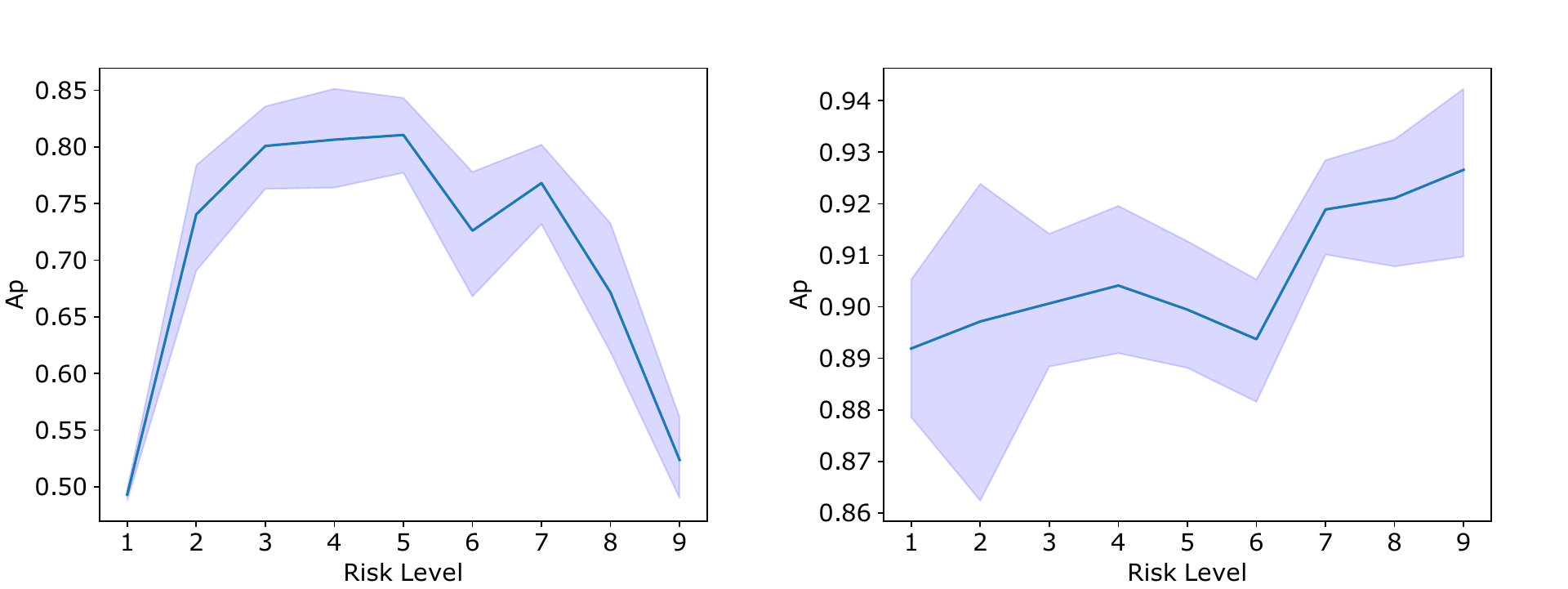}
\end{center}
   \caption{Average Precision as a function of the risk level. MNIST is on the left, and MHAD on the right. 95\% confidence intervals are shown in light blue. The baseline method, which does not leverage unlabeled data, is for risk $N=1$.}
\label{fig:ap_mnist_mhad}
\end{figure}

Similarly to the CIFAR experiments, we perform experiments with increasing risk levels $N$.
For each risk level, the experiment is repeated 10 times using different random initializations of the network's weights. This results in a total of 9 risk levels times 10 runs equals 90 experiments.
Figure \ref{fig:ap_mnist_mhad} shows the average precision for each risk level. Overall, risk-tolerant labeling outperforms conservative labeling($N= 1$) by up to 30 points of mean average precision with risk level $N=5$.
The impact of $N$ on the classification performance in Figure \ref{fig:ap_mnist_mhad} can be compared to the model with parameters $\alpha_{1} =3, \beta_{1}=1, \alpha_{2} =3$, and $\beta_{2}=9$ in Figure \ref{fig:model}. This indicates a similarly sudden increase of both positive and negative impacts, with an early positive impact and late negative impact. In that case, $N$ can be safely chosen in $[1,9]$ with little to no significant impact on the classification performance. 

\section{Experiments on Human Action Videos}
We tackle a more practical task by creating another sequential dataset based on video data from the Berkeley Multimodal Human Action Database (MHAD) (\url{https://tele-immersion.citris-uc.org/berkeley_mhad}). We select two actions, clapping hands (positive class) and jumping jacks (negative class), to create sequences of 10 consecutive frames. Each sequence starts with consecutive clapping frames and is switched at random to jumping jacks for the rest of the sequence (Figure~\ref{fig:image_sequences}).

We repeat the experimental setup of CIFAR and MNIST: experiments with 9 risks level $N$ and 10 repetition for each risk level. Figure \ref{fig:ap_mnist_mhad} shows the average precision for each risk level. Overall, risk-tolerant labeling outperforms conservative labeling($N= 1$) by up 30 points of mean average precision with risk level $N=5$. In contrast to the CIFAR and MNSIT experiments, for the MHAD experiments, the average precision continues to increase for risk level above $N=5$. This could be attributed to the high correlation between consecutive images. The impact of $N$ on the classification performance in Figure \ref{fig:precision} can be compared to the model with parameters $\alpha_{1} =0.5, \beta_{1}=5, \alpha_{2} =1$, and $\beta_{2}=25$ in Figure \ref{fig:model}, , with a negative impact substantially delayed. In that case, the risk level should be sampled larger than the average expected sequence length $M=5$. 

In addition to MHAD, we also show results on another huamn action video dataset,  HMDB51 \cite{Kuehne11}, in supplementary materials.

\section{Experiments on Videos of Hospital Patients}
We evaluate our method on a real-world task: events detection from continuous video recordings of hospital patients, using a dataset curated by our institute for this specific project. We present results comparing two levels of risk varying from $N=16$ (one 4 seconds clip at 4 FPS), the conservative model, to $N=10,000$, the risk-tolerant models, where $N$ is the number of frame following the sparse labels. We find that our risk-tolerant models significantly outperform the conservative model by 17 points of average precision.

\paragraph{\textbf{Dataset.}}
We aggregated and curated a dataset of continuous video recordings of hospital patients in an epilepsy center unit for children and neonates, collected with IRB oversight and approval.  
The dataset includes recordings from 59 neonate epileptic patients as part of the clinical routine of a hospital. 
The recording time per video lasts between 99 seconds and 720 minutes with a median of 33 minutes. To reduce training and inference times, the videos are downsampled from 25 to 4 frames per second (FPS), and the frame resolution is downsampled from 320x240 to 80x80.

As part of the clinical routine, these video recordings are sparsely labeled by multiple clinicians to indicate the occurrence of events with the corresponding start time only.
We identified 158 of those sparse labels indicating five types of events: patting of the neonates by nurses, suctioning of neonates' mouth liquid by nurses, rocking of neonates, neonate patient chewing food, and finally other types of cares being done on the patient by nurses, including changing intravenous line for example. Those events are selected as they can mislead automated seizure detection systems. 
The set of sparse labels is split randomly into training, validation and testing sets. Each set is class-balanced and data of different patients is used for each sets. This sampling leads to 30 sparse labels for training, 20 for validation and 25 for independent testing.
Videos in the testing set are fully reviewed to identify the end of the event. Non-overlapping 4-seconds long clips are sampled between the start and end of the event, and the task is defined as a five-classes video clip classification problem. 
For training and validation, we extract the $N$ frames following a sparse label as positive samples for that class, where $N$ is the risk level describe earlier. If the video is shorter, we stop at the end. Subsequently, 4-seconds (16 frames at 4 FPS) long non-overlapping clips are sampled in that range and given to the network.
More statistics about the dataset can be found in appendix.

\paragraph{\textbf{Training.}}
The networks are trained using a cross-entropy loss function and stochastic gradient descent optimizer with a learning rate of 0.001 and a momentum of 0.9. The networks are regularized with data augmentation including random color jitters with brightness, contrast, and saturation up to 0.8, and hue up to 0.4, random crops of 90 percent of the image size in x and y, and random horizontal flips. Models are training for 20 epochs. The best model is selected as the one with minimizing the cross-entropy on the validation set. 

\paragraph{\textbf{Results.}}

\begin{figure*}[t!]
\centering
\includegraphics[width=14cm]{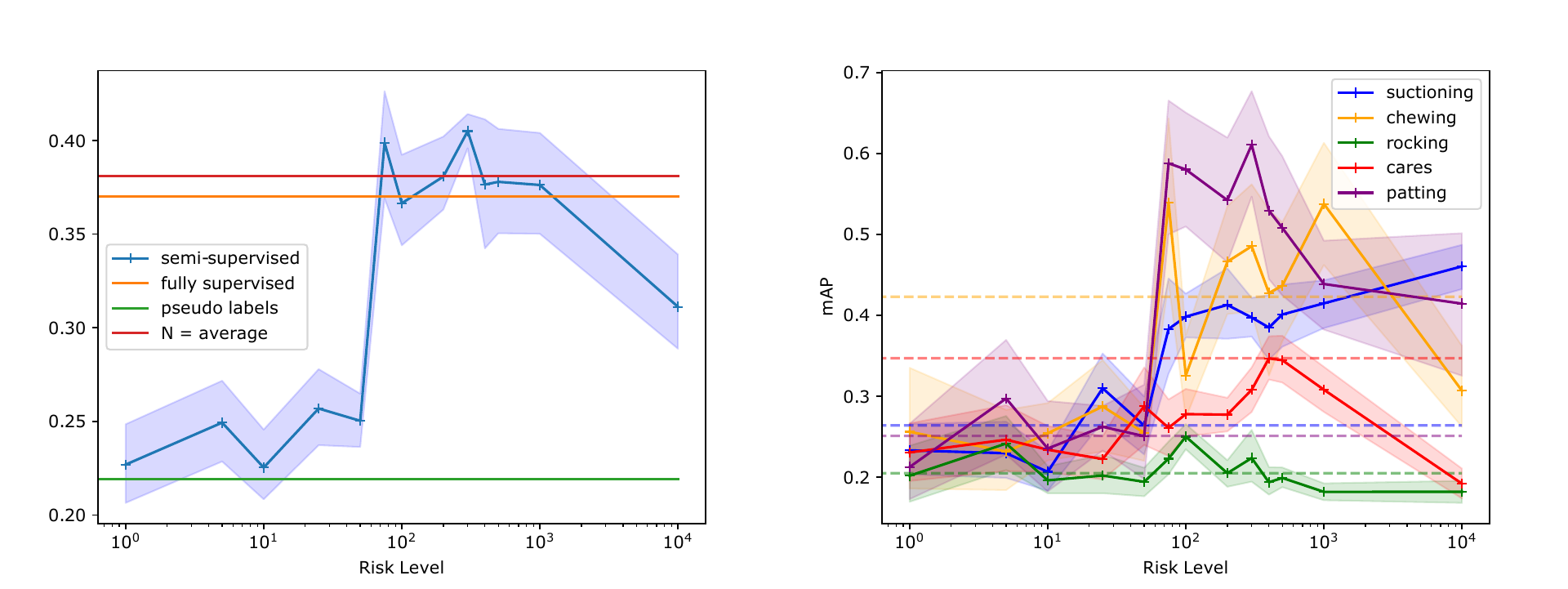}
\caption{\textbf{Left:} Mean Average Precision on the hospital video dataset under varying risk level $N$. Each semi-supervised point is averaged over 10 runs. We also plot the fully supervised reference, the pseudo label baseline and highlight results if the risk level $N$ was chosen as the average event time on the dataset. The pseudo labels results are averaged over 10 training runs. 95\% confidence intervals were computed using bootstrapping. 
\textbf{Right:} Class-wise Average Precision on the hospital video dataset. Dashed lines highlight results if the risk level $N$ was chosen as the class-wise average event time on the dataset.
}
\label{fig:mAP_video_all_classes}
\end{figure*}

Similarly to the experiments on the other datasets, we vary the risk level $N$, and for each risk level, the experiment is repeated 10 times using different random initializations. Figure \ref{fig:mAP_video_all_classes} shows the mean average precision for each risk level. Overall, risk-tolerant labeling significantly outperforms conservative labeling by up to 17 points of mean average precision with risk level $N=75$, which, at 4FPS, corresponds to the median duration of an event in our dataset (24 seconds). The evolution of classification performance with $N$ corresponds to the model with parameters $\alpha_{1} =5, \beta_{1}=5, \alpha_{2} =1$, and $\beta_{2}=10$ in Figure \ref{fig:model}. This indicates a late impact of both positive and negative sigmoid in Equation \ref{eq:perf}, with a sudden increase for the positive impact and a slower increase for negative impact. For comparison, we also reviewed training and validation video clips to locate the end of events and enable fully supervised training. For values of $N$ in $[100,1000]$, the proposed semi-supervised approach reached a classification performance similar to that of fully supervised training.

In addition, we compare risk-tolerant models with two other approaches: setting the risk level as the average length of events in our datasets, and using pseudo labeling \cite{lee2013pseudo} (Figure \ref{fig:mAP_video_all_classes}). Pseudo labeling reaches a performance statistically similar to that of the conservative models ($N=1$). By using the average length of events, we get close to the best performance for $N=300$, yet the results of the proposed risk-tolerant model are significantly higher (bootstrapped 95\% CI). Class-wise results (Figure \ref{fig:mAP_video_all_classes}) show that, while using the average class-wise event length allows some of the classes to reach their top accuracy (cares), other classes do not follow this pattern (suctioning, patting).

\paragraph{\textbf{Merging Classes: Suctioning.}}

Figure \ref{fig:mAP_video_all_classes} shows the class-wise mean average precision. Contrary to the other classes, the classification performance for the suctioning class does not decrease after $N=1000$. The performance even slightly increases. This is counter-intuitive as the median duration of suctioning events is 10 seconds with an average 14.36 seconds and a standard deviation of 14.36 seconds. At 4FPS, this would mean that most of those events last between no longer 120 frames, which is very far from $N=1000$. It appears that suctioning events most frequently occur as part of a longer event, which we can call "emergency event", which we did not account for in our classification problem. This emergency event class is part of the outlier class $C_{o}$ (method section).

To quantify this, for $N=1000$, we assessed the visual similarity between frames sampled as positives for suctioning versus that of other classes using the Structural Similarity Index Measure (SSIM) \cite{taghanaki2021}. We found that the SSIM between two random frames sampled as suctioning was significantly smaller (p-value $<$ 0.5) than a random suctioning frame and a random frame from any of the other 4 classes. On the contrary, repeating this experience for each of the other classes (excluding suctioning) did not indicate any significant difference in SSIM. We can conclude that $|X_{m,c} \cap X_{C_{o},c}|$ is close to $1$ and $P_{multi,c}$ close to $0$ in Equation \ref{eq:overlap}. Consequently, for the suctioning class, the impact of $N$ on the classification performance in Equation \ref{eq:perf} can be approximated to a single sigmoid describing the positive impact of $N$ as in Equation \ref{eq:single_sig}, with a $\beta_{2}$ similar to that of other classes, close to $N=100$. That is the trend that we observe for suctioning in Figure \ref{fig:mAP_video_all_classes}. This correspond to parameters $\alpha_{1} =0.5, \beta_{1}=5, \alpha_{2} =1$, and $\beta_{2}=25$ in Figure \ref{fig:model}, with a negative impact substantially delayed.

To verify the hypothesis qualitatively, we also inspected attention maps corresponding to correct prediction of the suctioning class. Guided-backpropagation attention maps \cite{springenberg2014} indicated that for most clips, the models focus of surrounding scene elements indicating the commotion instead of the suctioning device itself. On the contrary, for example for chewing, the models focused on the food, plate and hands of the patients; for patting, the models focused on the hand of the caregiver.

\section{Discussion and Limitations}

Choosing a risk level $N$ that was too high never significantly worsened the classification performance w.r.t that of the conservative models. In some datasets (e.g. MNIST), a wide range of risk-levels enabled reaching the optimal classification performance. On others (hospital video dataset), $N$ had to be chosen above a certain threshold. Finally, some datasets (CIFAR) had a more restrictive range of values of $N$ that allowed reaching the optimal classification performance. We proposed a model that could measure these patterns using four parameters $\alpha_{1}, \beta_{1}, \alpha_{2}$, and $\beta_{2}$. Correct understanding of the target classes and data could allow choosing these parameter prior to training and anticipate the optimal value for the risk level $N$.  We also give insight into estimating those parameters in case of merging classes. For example in Sect. 7.4, using SSIM, we identify that the \textit{suctioning} class is a subclass of a hidden class \textit{emergency events}, and conclude that $\alpha_{2}$ and $\beta_{2}$ can be set to values that minimize the negative impact of increasing $N$. 

 Most standard semi-supervised methods, including standard pseudo labels, \cite{lee2013pseudo} disregard the sequentiality of labels by considering samples i.i.d. and consequently fail to reach a satisfying performance on our real-world video dataset (22 mAP vs. 40 mAP for the proposed approach). We leverage the sequentiality of the data by estimating event end times from start times and discuss the quantified results of our proposed method in Figure 6. 

One of the limitations of the proposed model is that the sparse labels $l$ are always considered to be correct. In our experiments with continuous video recordings, review of data outside the curated dataset, revealed that some of the sparse labels are incorrect: the labeled event was not found in the video. To improve the model of the influence of $N$ on the classification accuracy, one could incorporate a term in Equation \ref{eq:perf} that accounts for the risk of the sparse labels of being incorrect. 

\section{Conclusions}
We proposed a semi-supervised training strategy for sparsely-labeled sequential data and showed that it can improve detection performance over several baseline models. Convolutional neural networks were able to leverage the additional--yet partially incorrectly labeled--positive training samples to significantly improve the mean average precision. We demonstrated this improvement on image sequences from benchmark datasets and detection of video events, where the proposed approach reached a classification performance statistically similar to that of a fully supervised approach.

{\small
\bibliographystyle{ieee_fullname}
\bibliography{egbib}
}

\end{document}